# Knowledge Graph Embeddings: A Comprehensive Survey on Capturing Relation Properties

Guanglin Niu

Beihang University, School of Artificial Intelligence, beihangngl@buaa.edu.cn

Knowledge Graph Embedding (KGE) techniques play a pivotal role in transforming symbolic Knowledge Graphs (KGs) into numerical representations, thereby enhancing various deep learning models for knowledge-augmented applications. Unlike entities, relations in KGs are the carriers of semantic meaning, and their accurate modeling is crucial for the performance of KGE models. Firstly, we address the complex mapping properties inherent in relations, such as one-to-one, one-to-many, many-to-one, and many-to-many mappings. We provide a comprehensive summary of relation-aware mapping-based models, models that utilize specific representation spaces, tensor decomposition-based models, and neural network-based models. Next, focusing on capturing various relation patterns like symmetry, asymmetry, inversion, and composition, we review models that employ modified tensor decomposition, those based on modified relation-aware mappings, and those that leverage rotation operations. Subsequently, considering the implicit hierarchical relations among entities, we introduce models that incorporate auxiliary information, models based on hyperbolic spaces, and those that utilize the polar coordinate system. Finally, in response to more complex scenarios such as sparse and dynamic KGs, this paper discusses potential future research directions. We explore innovative ideas such as integrating multimodal information into KGE, enhancing relation pattern modeling with rules, and developing models to capture relation characteristics in dynamic KGE settings.

## 1 INTRODUCTION

Knowledge graphs (KGs) facilitate the establishment of diverse relations between entities through a directed graph structure, thereby endowing machines with capabilities akin to human understanding, inference, and application of common sense or domain-specific knowledge. While symbolic knowledge is highly interpretable for humans, it poses challenges for efficient machine processing. Drawing inspiration from word embedding techniques that convert symbolic words into numerical vectors, Knowledge Graph Embedding (KGE) models endeavor to embed symbolic entities and relations from KGs into a numerical representation space, preserving the original semantic and structural information of the KG [1-3]. This transformation enables efficient knowledge retrieval and reasoning through numerical computation. KGE techniques have been widely applied in relation extraction [4] during KG construction, entity alignment [5], KG completion [6], and in knowledge-enhanced downstream tasks such as question-answering systems [7], recommendation systems [8], and pre-trained language models [9]. As a result, KGE has emerged as a foundational and mainstream model for constructing and applying KGs [10].

KGE approaches primarily concentrate on learning embeddings of entities and relations in a numerical space and evaluating the plausibility of each piece of knowledge through scoring functions. Consequently, many KGE models focus on designing the representation space [11], the scoring functions [12], and the encoding mechanisms for entities and relations [13]. Although there are several review papers on KGE, they mainly offer comprehensive overviews of the current research status [14-20]. The semantics of KGs are predominantly reflected in the relations between entities, which exhibit complex mapping characteristics such as one-to-one, one-to-many, many-to-one, and many-to-many, as well as symmetrical, anti-symmetrical, inverse, and composite relation patterns. Furthermore, implicit hierarchical relations often exist between entities in KGs. The ability to accurately model these relation characteristics significantly impacts the effectiveness of KGE and its performance across various tasks. However, the existing literature lacks a comprehensive summary and review of KGE models from the perspective of modeling relation characteristics. This paper, therefore, organizes KGE models from the perspectives of complex mapping characteristics, various relation patterns, and hierarchical relations between entities, and discusses future development directions of related research.

This paper begins with an overview of the basic principles of KGE models and several types of relation characteristics in Section 2. It then discusses KGE models for modeling complex relations in Section 3. KGE models for modeling various relation patterns are introduced in Section 4. Section 5 summarizes KGE models for modeling hierarchical relations between entities. Finally, Section 6 clarifies the future research directions of KGE.

## 2 INTRODUCTION OF KG EMBEDDING

KGs are instrumental in structuring knowledge as a directed graph, where nodes denote entities and edges represent the relationships between them. This framework enables machines to process and reason with human-like understanding of knowledge. However, the symbolic representation of knowledge, while interpretable by humans, poses computational challenges for large-scale KGs and their integration into deep learning models, such as pre-trained language models.

Knowledge Graph Embedding (KGE) techniques have emerged as a solution to this challenge. They automatically learn numerical vector representations of entities and relations from KG triples, thereby preserving the semantic and structural integrity of the KG in a form that is amenable to computational processes [22]. During the training phase, KGE models focus on establishing a numerical representation space for entities and relations, alongside a scoring function to evaluate the likelihood of triples. The objective function is meticulously designed to train the triples samples, allowing the model to learn embeddings that capture the essence of the KG [23]. Post-training, these embeddings can be deployed in various downstream applications, including the enhancement of pre-trained language models and the completion of KGs [24]. Figure 1 depicts the canonical KGE model, TransE, which translates symbolic KGs into numerical representations. The central premise of TransE is that the vector representations of a relation r and its associated entities h and t in a valid triple should adhere to the translational constraint $\mathbf{h} + \mathbf{r} = \mathbf{t}$. This constraint ensures that the learned embeddings maintain the original semantic and structural properties of the KG [25]. TransE has garnered significant attention in the KGE domain due to its simplicity and effectiveness, with numerous studies extending its foundational principles.

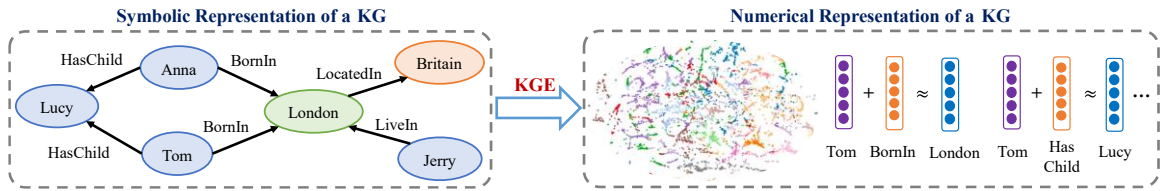

Figure 1: Example diagram of KGE technique

The efficacy of KGE models is typically assessed using a link prediction evaluation framework. This method determines the model's proficiency in predicting new triples after the completion of training. For instance, when predicting triples with a missing head entity (?, relation, tail entity), the process involves constructing candidate triples by substituting all entities in the KG for the head entity position and subsequently calculating the rank of the correct candidate triple among all contenders. Three prevalent evaluation metrics for KGE tasks include:

(1) Mean Rank (MR), which indicates the average rank of all correct candidate triples.

(2) Mean Reciprocal Rank (MRR), which represents the average of the reciprocal ranks of all correct triples.

(3) Hits@N, which signifies the proportion of correct candidate triples ranked within the top N positions. A lower MR value and higher MRR or Hits@N values are indicative of superior KGE model performance.

The semantic richness of KGs is predominantly conveyed through the relations that interlink entities. These relations exhibit a variety of characteristics, including complex mapping features, diverse relation patterns, and hierarchical structures among entities. A pivotal challenge in contemporary KGE research is the effective modeling of these relation characteristics within the numerical representation space, which is vital for the performance of KGE. For enhanced clarity, a detailed explanation of several relation properties is provided as followings.

## 2.1 Introduction of Complex Mapping Features of Relations

Complex mapping features of relations include one-to-one (1-1), one-to-many (1-N), many-to-one (N-1), and many-to-many (N-N) relations. In this context, the entity in the *1* position is unique, implying that when the relation and another entity in a triple are determined, substituting the unique entity with any other entity results in incorrect knowledge. Conversely, the entity in the N position is non-unique, suggesting multiple valid choices for this position. For instance, in the N-1 relation *London BornIn*, the tail entity *London* typically corresponds to multiple head entities, while a specific head entity, such as *Tom*, only corresponds to a single tail entity. Figure 2 illustrates instances of complex mapping features of relations.

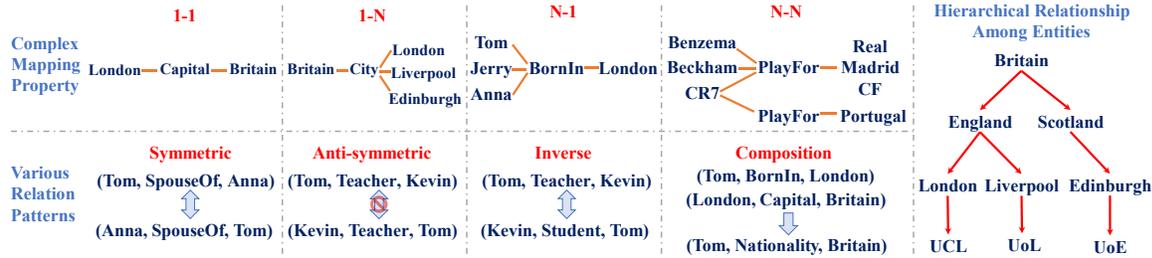

Figure 2: Example diagram of several relation properties

## 2.2 Introduction of Diverse Relation Patterns

Diverse relation patterns encompass symmetric, asymmetric, inverse, and composite relations. Symmetric relations, such as *Classmate* or *SpouseOf*, and asymmetric relations, such as *Teacher*, are inherent characteristics of the relations themselves. Triples containing symmetric relations remain valid even when the head and tail entities are interchanged. In contrast, triples with asymmetric relations become invalid upon swapping the head and tail entities. The inverse relation pattern involves two distinct relations that are inverses of each other. For instance, the triples (*Tom*, *Teacher*, *Kevin*) and (*Kevin*, *Student*, *Tom*) are both valid. Composite relations involve three relations where two relations can be combined to form another relation. For example, based on the two factual triples (*Tom*, *BornIn*, *London*) and (*London*, *Capital*, *Britain*), one can directly infer (*Tom*, *Nationality*, *Britain*), as depicted in Figure 2. To ascertain whether a KGE model can effectively model a specific relation pattern, it is essential to verify if the triples representing the premises of the relation pattern are satisfied and align with the model's optimization objective. If the triple expressing the inference also aligns with the optimization objective, it indicates that the model can effectively model that type of relation pattern.

## 2.3 Introduction of Hierarchical Relations Between Entities

Since the two entities connected by each relation may belong to different semantic hierarchies, there are inherent hierarchical relations between entities. As illustrated in Figure 2, entities such as *Britain*, *England*, *London*, and *UCL* exhibit hierarchical relations. Modeling these hierarchical relations between entities is crucial for KGE models to effectively represent richer semantic information between entities. Due to the constraint goal of the TransE model being $\mathbf{h} + \mathbf{r} = \mathbf{t}$ for triples containing 1-N relations, such as (*Britain*, *City*, *London*) and (*Britain*, *City*, *Liverpool*), the vector embeddings of entities London and Liverpool will be forced to be approximately the same, which is clearly unreasonable. Similarly, for two triples containing symmetric relations, such as $(h, r, t)$ and $(t, r, h)$ where $h$, $r$ and $t$ represent the head entity, relation, and tail entity, respectively, TransE would result in $\mathbf{r} = \mathbf{0}$, rendering it incapable of effectively modeling symmetric relations. Furthermore, without auxiliary information, TransE cannot model implicit hierarchical relations between entities.

The limitation of TransE in modeling these relation characteristics stems from its use of translation operations in real vector space to represent relations between entities. Consequently, subsequent researches have focused on the modification of TransE. These approaches aim to model the complex mapping features of relations, various relation patterns, and hierarchical relations between entities. Therefore, this paper systematically reviews and summarizes the current research status concerning these three types of relation characteristics.

## 3    MODELS FOR COMPLEX RELATION MAPPING

The TransE model, while groundbreaking, faces limitations in capturing the nuances of complex relation mappings. To overcome these, current KGE models have adopted several innovative strategies, including relation-aware mapping of entity vector embeddings, embedding KGs into specific geometric spaces, utilizing tensor decomposition techniques on high-dimensional third-order tensors representing KGs, and employing neural networks to learn the intricate interactions between entities and relations.

### 3.1    Models Based on Relation-Aware Mapping

A subset of relation-aware mapping models builds upon the translational operation core to TransE. These models introduce relation-dependent entity mapping mechanisms to achieve complex relation mappings. TransH [25] pioneered this approach by associating each relation with a unique hyperplane and projecting entities onto these relation-specific hyperplanes. This innovation allows entities to possess distinct representations across different relations. The plausibility of a triple (h, r, t) is evaluated using the scoring function:

$$E(h, r, t) = \|\boldsymbol{h_r} + \boldsymbol{r} - \boldsymbol{t_r}\|_{L_1/L_2} \tag{1}$$

$$\boldsymbol{h_r} = \boldsymbol{h} - \boldsymbol{w}^T \boldsymbol{h} \boldsymbol{w}, \ \boldsymbol{t_r} = \boldsymbol{t} - \boldsymbol{w}^T \boldsymbol{t} \boldsymbol{w} \tag{2}$$

where $\boldsymbol{h} \in \mathbb{R}^k$, $\boldsymbol{r} \in \mathbb{R}^k$, $\boldsymbol{t} \in \mathbb{R}^k$ represent the embeddings of the head entity $h$, relation $r$, and tail entity $t$ in a k-dimensional real vector space. $\boldsymbol{w} \in \mathbb{R}^k$ is the normal vector of the hyperplane corresponding to relation $r$. $\boldsymbol{h_r} \in \mathbb{R}^k$ and $\boldsymbol{t_r} \in \mathbb{R}^k$ are the vector embeddings of the head and tail entities projected onto the hyperplane of relation $r$. $L_1/L_2$ represents the scalar score of the triple obtained through the $L_1$ or $/L_2$ norm.

From a different perspective, for one-to-many relations, such as the *City* relation where the same head entity, for example, *Britain*, corresponds to multiple tail entities, projecting these varied tail entity vectors onto the relation-specific hyperplane for *City* results in similar vector embeddings. This mechanism adeptly addresses the modeling challenges of one-to-many, many-to-one, and many-to-many complex mapping relations. However, TransH's requirement for all entities associated with the same relation to be projected onto a single hyperplane restricts entity expressiveness. TransR [26] addresses this by defining a relation space for each relation, projecting entities into specific relation spaces, which offers two representational learning advantages: (1) For a 1-N relation, multiple tail entities corresponding to the same head entity have distinct vector embeddings in entity space but similar embeddings in relation space; (2) Multiple originally similar entity vector embeddings can be effectively differentiated in relation space. The scoring function for TransR is defined as:

$$E(h, r, t) = \|\boldsymbol{M_r} \boldsymbol{h} + \boldsymbol{r} - \boldsymbol{M_r} \boldsymbol{t}\|_{L_1/L_2} \tag{3}$$

where $\boldsymbol{M_r} \in \mathbb{R}^{d \times k}$ is the projection matrix for relation $\boldsymbol{h} \in \mathbb{R}^k$ and $\boldsymbol{t} \in \mathbb{R}^k$ are the vector embeddings of the head and tail entities, and $\boldsymbol{r} \in \mathbb{R}^d$ is the vector embedding of the relation.

TransR's projection matrix is solely related to the relation, lacking the influence of the semantic information of the head and tail entities on the projection process. STransE [27] enhances TransR by learning two separate projection matrices for each relation, enabling the head and tail entities to be projected into their respective relational spaces. The scoring function is defined as:

$$E(h, r, t) = \|\boldsymbol{M_{r1}} \boldsymbol{h} + \boldsymbol{r} - \boldsymbol{M_{r2}} \boldsymbol{t}\|_{L_1/L_2} \tag{4}$$

where $\boldsymbol{M_{r1}} \in \mathbb{R}^{k \times k}$ and $\boldsymbol{M_{r2}} \in \mathbb{R}^{k \times k}$ are the two projection matrices for relation $r$.

However, STransE faces high complexity due to its extensive parameter set. To mitigate this, TransD [28] introduces a dynamic projection matrix constructed by integrating the relation projection vector, the head entity projection vector, and the tail entity projection vector. This approach enables the head and tail entities in a factual triple to be mapped through distinct projection matrices. The scoring function is defined as follows:

$$E(h, r, t) = \|\boldsymbol{M}_{rh}\boldsymbol{h} + \boldsymbol{r} - \boldsymbol{M}_{rt}\boldsymbol{t}\|_{L_1/L_2} \tag{5}$$

$$\boldsymbol{M}_{rh} = \boldsymbol{w}_r \boldsymbol{w}_h^T + \boldsymbol{I}^{d \times k}, \quad \boldsymbol{M}_{rt} = \boldsymbol{w}_r \boldsymbol{w}_t^T + \boldsymbol{I}^{k \times d} \tag{6}$$

in which $\boldsymbol{h} \in \mathbb{R}^k$ and $\boldsymbol{t} \in \mathbb{R}^k$ represent the vector embeddings of the head entity $h$ and the tail entity $t$ respectively. $\boldsymbol{w}_h \in \mathbb{R}^k$ and $\boldsymbol{w}_t \in \mathbb{R}^k$ are the projection vectors related to the head and tail entities, and $\boldsymbol{w}_r \in \mathbb{R}^d$ is the relation projection vector. The creation of specific projection matrices for the head and tail entities using their respective projection vectors, in conjunction with the relation projection vector, significantly reduces the number of parameters compared to the STransE model, which directly employs two projection matrices.

Similarly, TranSparse [29] calculates sparsity by assessing the number of entity pairs connected by each relation, constructing adaptive projection sparse matrices. This method addresses the imbalance of relations in KGs by projecting the head and tail entities adaptively. The scoring function is defined as follows:

$$E(h, r, t) = \|\theta_h \boldsymbol{M}_{rh}\boldsymbol{h} + \boldsymbol{r} - \theta_h \boldsymbol{M}_{rt}\boldsymbol{t}\|_{L_1/L_2} \tag{7}$$

$$\theta_h = 1 - (1 - \theta_{min})N_{rh}/N^* \tag{8}$$

$$\theta_t = 1 - (1 - \theta_{min})N_{rt}/N^* \tag{9}$$

where $\theta_h$ and $\theta_t$ represent the sparsity of the head entity projection matrix and tail entity projection matrix, respectively. $N_{rh}$ and $N_{rt}$ are the numbers of head and tail entities associated with relation $r$, and $N^*$ is the maximum of $N_{rh}$ and $N_{rt}$. $\theta_{min} \in [0,1]$ is a hyperparameter that controls sparsity.

TransF [30] relaxes the optimization objective in TransE to ensure that the vector resulting from $\boldsymbol{h} + \boldsymbol{r}$ only needs to maintain the same direction as the tail entity vector embedding. Thus, for multiple tail entities in a 1-N relation, different vector embeddings can be learned for each tail entity, effectively modeling the complex mapping characteristics of relations. The scoring function is defined as follows:

$$E(h, r, t) = (\boldsymbol{h} + \boldsymbol{r})^T \boldsymbol{t} + \boldsymbol{h}^T(\boldsymbol{t} - \boldsymbol{r}) \tag{10}$$

Contrary to evaluation mechanisms that use Euclidean distance to calculate the plausibility of triples, TransA [31] represents each relation as a symmetric non-negative matrix and employs Mahalanobis distance in the scoring function to measure the plausibility of triples. This approach treats the relation matrix as the weighted sum of the difference vectors between head, tail entities, and relation vectors across all dimensions. This enables multiple tail entities in a 1-N relation to have different vector embeddings. Here, multiple triples formed by the same head entity and relation with different tail entities can satisfy similar adaptive Mahalanobis distance values, thereby modeling the complex mapping characteristics of relations. The scoring function is defined as follows:

$$E(h, r, t) = (|\boldsymbol{h} + \boldsymbol{r} - \boldsymbol{t}|)^T \boldsymbol{M}_r(|\boldsymbol{h} + \boldsymbol{r} - \boldsymbol{t}|) \tag{11}$$

where $(|\boldsymbol{h} + \boldsymbol{r} - \boldsymbol{t}|) \in \mathbb{R}^k$ can be shown as $(|h_1 + r_1 - t_1|, |h_2 + r_2 - t_2|, \cdots, |h_k + r_k - t_k|)$, and $\boldsymbol{M}_r \in \mathbb{R}^{k \times k}$ indicates the symmetric non-negative matrix corresponding to relation $r$.

TransM [32] assigns different triple scoring weights for different relations. Relations with complex mapping characteristics are given lower weights. For instance, multiple triples containing the same head entity and 1-N relation but different tail entities can have different vector embeddings. According to TransE, these triples would only achieve different score values. However, the TransM scoring function can ensure that these triples have similar scores after adjusting for lower weights. The scoring function is defined as follows:

$$E(h, r, t) = w_r \|\boldsymbol{h} + \boldsymbol{r} - \boldsymbol{t}\|_{L_1/L_2} \tag{12}$$

where $w_r$ is the triple scoring weight for relation $r$.

To facilitate understanding, taking the modeling of the complex mapping characteristics of 1-N relations as an example, the core ideas of several typical models based on relation-aware mapping are illustrated in Figure 3.

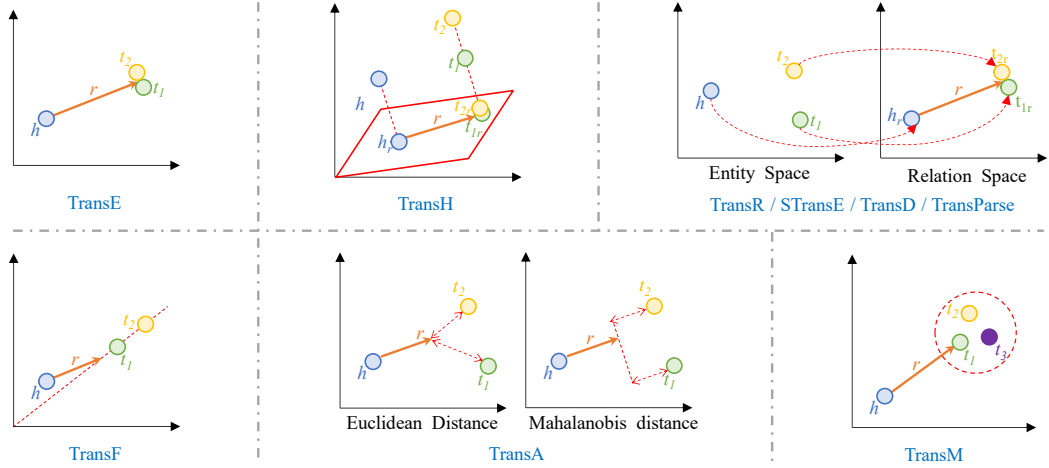

Figure 3: Illustrative diagram of the 1-N relationship modeling utilizing relation-aware mapping strategies

## 3.2 Models Based on Specific Representation Spaces

While models based on relation-aware mapping embed KGs into Euclidean vector spaces, models based on specific representation spaces embed KGs into diverse spaces such as Gaussian, manifold, and Lie groups. These specialized representation spaces inherently address the modeling challenges of complex relation characteristics.

**KG2E** [33] embeds entities and relations into a high-dimensional Gaussian space, represented by Gaussian distributions: $\boldsymbol{h} \sim N(\boldsymbol{u}_h, \Sigma_h)$, $\boldsymbol{r} \sim N(\boldsymbol{u}_r, \Sigma_r)$, $\boldsymbol{t} \sim N(\boldsymbol{u}_t, \Sigma_t)$ where the mean vectors $\boldsymbol{u}_h \in \mathbb{R}^k$, $\boldsymbol{u}_r \in \mathbb{R}^k$ and $\boldsymbol{u}_t \in \mathbb{R}^k$ represent the central positions of entities and relations in the representation space, and the covariance matrices $\Sigma_h \in \mathbb{R}^{k \times k}$, $\Sigma_r \in \mathbb{R}^{k \times k}$, $\Sigma_t \in \mathbb{R}^{k \times k}$ represent the uncertainties of entities and relations. KG2E evaluates the plausibility of a triple by computing the similarity between the difference of entity pairs $(\boldsymbol{t} - \boldsymbol{h}) \sim N(\boldsymbol{u}_e = \boldsymbol{u}_t - \boldsymbol{u}_h, \Sigma_e = \Sigma_t + \Sigma_h)$ and $r$. It employs KL divergence and expected likelihood to design the scoring function $E(h, r, t)$ as follows:

$$\int_{x \in \mathbb{R}^k} N(\boldsymbol{x}, \boldsymbol{u}_r, \Sigma_r) log \frac{N(\boldsymbol{x}, \boldsymbol{u}_e, \Sigma_e)}{N(\boldsymbol{x}, \boldsymbol{u}_r, \Sigma_r)} d\boldsymbol{x} \tag{13}$$

$$\int_{x \in \mathbb{R}^k} N(\boldsymbol{x}, \boldsymbol{u}_e, \Sigma_e) N(\boldsymbol{x}, \boldsymbol{u}_r, \Sigma_r) d\boldsymbol{x} \tag{14}$$

where for different tail entities in a 1-N relation, using these two similarity scoring functions can yield similar triple scores, thereby modeling complex relation characteristics.

ManifoldE [34] represents entities and relations in a manifold space, such as a high-dimensional sphere. It requires that the head entity and relation of each triple be the center of the sphere, with the tail entity positioned within the sphere. In this scenario, different tail entities in a 1-N relation only need to satisfy the condition of being within the sphere, cleverly achieving the effect of modeling complex relation characteristics. The scoring function is designed as follows:

$$E(h, r, t) = \|MF(\boldsymbol{h}, \boldsymbol{r}, \boldsymbol{t}) - D_r^2\| \tag{15}$$

where $MF$ is the manifold function, and $D_r$ is the manifold parameter representing the radius of the sphere.

TorusE [35] embeds KGs onto a compact Lie group torus. The characteristic of this representation space is that an entity representation $[\boldsymbol{h}] \in \mathrm{T}^k$ and two different entity representations $[\boldsymbol{t}_1] \in \mathrm{T}^k$ and $[\boldsymbol{t}_2] \in \mathrm{T}^k$ can have similar differences on the Lie group torus, i.e., $[\boldsymbol{t}_1] - [\boldsymbol{h}]$ and $[\boldsymbol{t}_2] - [\boldsymbol{h}]$. Adopting the optimization objective from TransE $[\boldsymbol{h}] + [\boldsymbol{r}] = [\boldsymbol{t}]$, these two differences $[\boldsymbol{t}_1] - [\boldsymbol{h}]$ and $[\boldsymbol{t}_2] - [\boldsymbol{h}]$ can represent the same relation embedding, thus modeling 1-N relations. TorusE defines three scoring functions as follows:

$$E_{L1}(h, r, t) = 2d_{L1}([\boldsymbol{h}] + [\boldsymbol{r}], [\boldsymbol{t}]) \tag{16}$$

$$E_{L2}(h, r, t) = (2d_{L2}([\boldsymbol{h}] + [\boldsymbol{r}], [\boldsymbol{t}]))^2 \tag{17}$$

$$E_{eL2}(h, r, t) = (d_{eL2}([\boldsymbol{h}] + [\boldsymbol{r}], [\boldsymbol{t}])/2)^2 \tag{18}$$

Several KGE models with various specific representation spaces are illustrated in Figure 4.

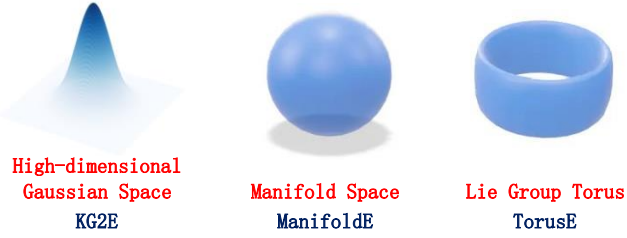

Figure 4: Illustrative diagram of the specific representation spaces

### 3.3 Tensor Decomposition-Based Models

A variety of KGE models that leverage tensor decomposition techniques initially conceptualize the KG as a large-scale third-order tensor. Within this tensor, each element's value indicates the presence or absence of a corresponding triple in the KG: a value of *1* denotes the existence of the triple, while a *0* signifies its non-existence. Through tensor decomposition, the score of each triple is transformed into a bilinear operation among the representations of the head entity, tail entity, and the relation. This operation inherently addresses the complexities of modeling various relation characteristics.

RESCAL [36] is the pioneering KGE model to employ tensor decomposition. It computes the score of each triple within the tensor through matrix multiplication involving low-dimensional entity vector embeddings and the relation matrix representations. The scoring function of RESCAL is defined as follows:

$$E(h, r, t) = \boldsymbol{h}^T \boldsymbol{M_r} \boldsymbol{t} = \sum_{i=0}^{k-1} \sum_{j=0}^{k-1} [\boldsymbol{M_r}]_{ij} [\boldsymbol{h}]_i [\boldsymbol{t}]_j \tag{19}$$

where $\boldsymbol{h} \in \mathbb{R}^k$ and $\boldsymbol{t} \in \mathbb{R}^k$ represent the entity embeddings of the head entity $h$ and the tail entity $t$ in a k-dimensional real vector space, while $\boldsymbol{M_r} \in \mathbb{R}^{k \times k}$ is the matrix representation of the relation $r$. Matrix multiplication facilitates obtaining analogous scores for different tail entity vectors associated with the same 1-N relation and head entity.

DistMult [37] simplifies RESCAL by utilizing a diagonal matrix for each relation representation, substantially reducing the number of parameters for each relation. The scoring function is designed as:

$$E(h, r, t) = \boldsymbol{h}^T diag(\boldsymbol{r}) \boldsymbol{t} = \sum_{i=0}^{k-1} [\boldsymbol{h}]_i [\boldsymbol{r}] [\boldsymbol{t}]_i \tag{20}$$

where $diag(\boldsymbol{r}) \in \mathbb{R}^{k \times k}$ denotes a diagonal matrix, and $\boldsymbol{r} \in \mathbb{R}^k$ is the vector embedding of the relation.

TuckER [38] employs Tucker decomposition, which decomposes a third-order tensor into a core tensor and three factor matrices. Based on this concept, TuckER calculates the score of each triple using tensor multiplication between the core tensor and the head entity vector, relation vector, and tail entity vector. The scoring function of TuckER is:

$$E(h, r, t) = w \times_1 \boldsymbol{h} \times_2 \boldsymbol{r} \times_3 \boldsymbol{t} \tag{21}$$

in which $w \in \mathbb{R}^{k \times d \times k}$ represents the core tensor, which can be considered shared weight parameters. $\boldsymbol{h} \in \mathbb{R}^k$ and $\boldsymbol{t} \in \mathbb{R}^k$ are the vector embeddings of the head and tail entities, respectively, and $\boldsymbol{r} \in \mathbb{R}^d$ is the vector embedding of the relation. The symbol $\times_n$ denotes the n-th mode tensor product.

### 3.4 Neural Network-Based Models

KGE models that employ neural networks utilize nonlinear operations and network architectures to capture the interactions between entities and relations, effectively addressing the complexity of modeling diverse relation characteristics.

SME [39] is among the pioneering approaches to implement KGE using neural networks. Within its fully connected neural network, the vector embeddings of the head and tail entities are combined with the relation vector embeddings in the hidden layers to produce two hidden layer vector embeddings. The inner product of these vectors is then calculated to score the triple. The scoring function of SME is defined as follows:

$$g_u(h, r) = \boldsymbol{M}_u^1 \boldsymbol{h} + \boldsymbol{M}_u^2 \boldsymbol{r} + \boldsymbol{b}_u \tag{22}$$

$$g_v(t, r) = \boldsymbol{M}_v^1 \boldsymbol{h} + \boldsymbol{M}_v^2 \boldsymbol{r} + \boldsymbol{b}_v \tag{23}$$

$$E(h, r, t) = g_u(h, r)^T g_v(t, r) \tag{24}$$

where $\boldsymbol{M}_u^1 \in \mathbb{R}^{d \times k}$, $\boldsymbol{M}_u^2 \in \mathbb{R}^{d \times k}$, $\boldsymbol{M}_v^1 \in \mathbb{R}^{d \times k}$ and $\boldsymbol{M}_v^2 \in \mathbb{R}^{d \times k}$ represent the weight matrices of the neural network. $\boldsymbol{b}_u \in \mathbb{R}^d$ and $\boldsymbol{b}_v \in \mathbb{R}^d$ represent the bias vectors. NTN [40] inputs the vector embeddings of the head and tail entities into a relation-specific neural network. The scoring of the triple is achieved through tensor multiplication, matrix multiplication, and activation functions. The scoring function for NTN is designed as follows:

$$E = \boldsymbol{r}^T \tanh\left(\boldsymbol{h}^T \boldsymbol{M}_r \boldsymbol{t} + \boldsymbol{M}_r^1 \boldsymbol{h} + \boldsymbol{M}_r^2 \boldsymbol{t} + \boldsymbol{b}_r\right) \tag{25}$$

where $\boldsymbol{h} \in \mathbb{R}^k$ and $\boldsymbol{t} \in \mathbb{R}^k$ are the vector embeddings of the head and tail entities, respectively. $\boldsymbol{r} \in \mathbb{R}^d$ is the vector embedding of the relation. $\boldsymbol{M}_r \in \mathbb{R}^{k \times k \times d}$, $\boldsymbol{M}_r^1 \in \mathbb{R}^{d \times k}$, and $\boldsymbol{M}_r^2 \in \mathbb{R}^{d \times k}$ are the bilinear and linear mapping parameters in the neural network, while $\boldsymbol{b}_r \in \mathbb{R}^d$ indicates the bias vector.

ConvE [41] and ConvKB [42] draw on techniques from computer vision where convolutional neural networks (CNNs) extract feature maps from 2D images. They reconstruct the vector embeddings of head and tail entities, as well as relations, into 2D matrices. Convolutional kernels are then used to encode interactions between entities and relations. The key difference between these two models is that ConvE reconstructs the head entity and relation embeddings into a 2D matrix, which after convolution operations, combines with the tail entity vector to obtain the triple score:

$$E(h, r, t) = \sigma(vec(\sigma([\boldsymbol{M}_h; \boldsymbol{M}_r] * \omega))\boldsymbol{W}))\boldsymbol{t} \tag{26}$$

in which $\boldsymbol{M}_h \in \mathbb{R}^{dw \times dh}$ and $\boldsymbol{M}_r \in \mathbb{R}^{dw \times dh}$ are the matrix representations of the head entity and the relation, respectively. $\boldsymbol{t} \in \mathbb{R}^k$ represents the tail entity vector, $\omega$ is the convolution kernel, $*$ denotes the convolution operation, $vec$ represents tensor reconstruction into a vector, and $\boldsymbol{W}$ is the linear mapping matrix.

In contrast, ConvKB stacks the head, tail entities, and relation vectors directly to construct a 2D matrix, utilizing convolution operations to extract the overall features of the triple. The scoring function of ConvKB is designed as follows:

$$E(h, r, t) = \text{concat}(Relu([\boldsymbol{h}; \boldsymbol{r}; \boldsymbol{t}] * \omega))\boldsymbol{w} \tag{27}$$

where $\boldsymbol{h} \in \mathbb{R}^k$ and $\boldsymbol{t} \in \mathbb{R}^k$ are the vector embeddings of the head and tail entities, respectively, $\boldsymbol{r} \in \mathbb{R}^d$ is the vector embedding of the relation, $Relu$ represents the rectified linear unit function, and $\boldsymbol{w}$ denotes the weight vector.

CapsE [43] is designed based on ConvKB by using convolution kernels to extract feature maps and employs a capsule network to capture the entries along the same dimension of the feature vectors. The scoring function of CapsE is defined as:

$$E(h, r, t) = \|capsnet(Relu([\boldsymbol{h}; \boldsymbol{r}; \boldsymbol{t}] * \omega))\| \tag{28}$$

where $capsnet$ represents the capsule network operation.

To enhance the interaction between entities and relations, InteractE [44] builds upon ConvE by employing more complex techniques such as direct stacking, row-wise circular convolution, and element-wise interaction. The embeddings are reconstructed into a 2D matrix using these models, and the scoring function is further designed as follows:

$$E(h, r, t) = \sigma(vec(Relu(([\boldsymbol{M}_h; \boldsymbol{M}_r] * \omega))\boldsymbol{W}))\boldsymbol{t} \tag{29}$$

A comparison of several neural network-based models in extracting interaction features between entities and relations is illustrated in Figure 5. To facilitate the comparison of models that handle complex relation mapping properties, Table 1 provides a summary and overview of these models.

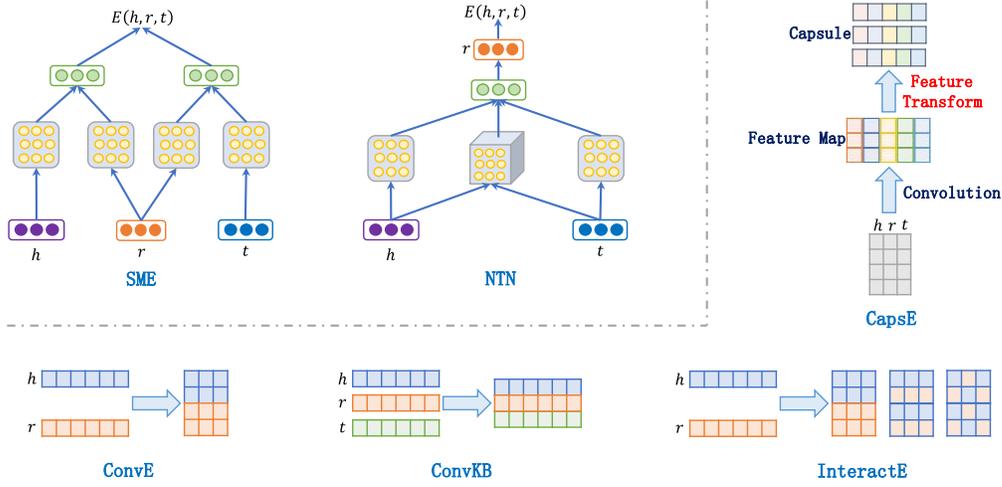

Figure 5: Illustrative diagram of the models based on neural networks

Table 1: A comprehensive summary of models that address complex mapping characteristics of relations within the context of KGE

| Model | | Embedding Space | Scoring Function | Characteristics | Pros and Cons |
|---|---|---|---|---|---|
| Models Based on Relation-Aware Mapping | TransH | Real Vector Space | $\|\boldsymbol{h_r} + \boldsymbol{r} - \boldsymbol{t_r}\|_{L_1/L_2}$ | Relation hyperplane projection | Pros: robust geometric interpretability. Cons: require additional parameters for each relation, which can lead to increased computational costs and potential overfitting. |
| | TransR | | $\|\boldsymbol{M_r}\boldsymbol{h} + \boldsymbol{r} - \boldsymbol{M_r}\boldsymbol{t}\|_{L_1/L_2}$ | Relation space projection | |
| | STransE | | $\|\boldsymbol{M_{r1}}\boldsymbol{h} + \boldsymbol{r} - \boldsymbol{M_{r2}}\boldsymbol{t}\|_{L_1/L_2}$ | Projection of Head and Tail Entities into Different Relation Spaces | |
| | TransD | | $\|\theta_h \boldsymbol{M_{rh}}\boldsymbol{h} + \boldsymbol{r} - \theta_h \boldsymbol{M_{rt}}\boldsymbol{t}\|_{L_1/L_2}$ | Adaptive Sparse Projection Matrix | |
| | TransF | | $(\boldsymbol{h} + \boldsymbol{r})^T \boldsymbol{t} + \boldsymbol{h}^T(\boldsymbol{t} - \boldsymbol{r})$ | "Head Entity Vector + Relation Vector" aligned with Tail Entity Vector | |
| | TransA | | $(|\boldsymbol{h} + \boldsymbol{r} - \boldsymbol{t}|)^T \boldsymbol{M_r}(|\boldsymbol{h} + \boldsymbol{r} - \boldsymbol{t}|)$ | Euclidean Distance replaced by Weighted Mahalanobis Distance | |
| | TransM | | $w_r \|\boldsymbol{h} + \boldsymbol{r} - \boldsymbol{t}\|_{L_1/L_2}$ | Reducing Weights for Relations with Complex Mapping Characteristics | |
| Models Based on Specific Representation Spaces | KG2E | Real Vector Space | $\int_{\boldsymbol{x} \in \mathbb{R}^k} N(\boldsymbol{x}, \boldsymbol{u_e}, \Sigma_e) N(\boldsymbol{x}, \boldsymbol{u_r}, \Sigma_r) d\boldsymbol{x}$ | Representing Uncertainty of Entities and Relations in Gaussian Space | Pros: directly modeling complex relation mappings. Cons: a more complex representation space compared to real vector spaces. |
| | ManifoldE | Real Vector Space | $\|MF(\boldsymbol{h}, \boldsymbol{r}, \boldsymbol{t}) - D_r^2\|$ | Head Entity and Relation as Center of Sphere, Tail Entity within the Sphere | |
| | TorusE | Real Vector Space | $E_{L_1}(h, r, t) = 2d_{L_1}([\boldsymbol{h}] + [\boldsymbol{r}], [\boldsymbol{t}])$ | Consistent Difference of Embeddings on Compact Lie Group Torus for Different Tail Entities | |
| Tensor Decomposition- | RESCAL | Real Vector Space | $\boldsymbol{h}^T \boldsymbol{M_r} \boldsymbol{t}$ | Decomposing third-order tensors into low-dimensional matrices and tensor multiplication | Pros: directly capture complex |

| Based Models | DistMult | | $h^T diag(r)t$ | Representing Each Relation as a Diagonal Matrix to Reduce Parameters | relation mappings. |
|---|---|---|---|---|---|
| | TuckER | | $w \times_1 h \times_2 r \times_3 t$ | Reducing Parameters through Shared Weight Mechanism | Cons: come with a higher parameter count. |
| Neural Network-Based Models | SME | Real Vector Space | $g_u(h,r)^T g_v(t,r)$ | Using Fully Connected Neural Networks to Model Interactions between Head, Tail Entities, and Relations for Triple Scoring | |
| | NTN | | $r^T \tanh(h^T M_r t + M_r^1 h + M_r^2 t + b_r)$ | Encoding Interactions between Head, Tail Entities, and Relations in Fully Connected Neural Networks | |
| | ConvE | | $\sigma(vec(\sigma(([M_h; M_r] * \omega))W))t$ | Reconstructing head entity and relation embeddings into 2D matrices for convolution operations | Pros: directly modeling complex relation mappings. |
| | ConvKB | | $concat(Relu([h; r; t] * \omega))w$ | Directly stacking head, tail entity, and relation vectors to form 2D matrices for convolution operations | Cons: high computational complexity. |
| | CapsE | | $\|capsnet(Relu([h; r; t] * \omega))\|$ | Using capsule networks post-convolution to capture entries along the same dimension of feature vectors | |
| | InteractE | | $\sigma(vec(Relu(([M_h; M_r] * \omega))W))t$ | Reconstructing 2D matrices for convolution operations using direct stacking, row-wise circular convolution, and element-wise interaction | |

## 4 MODELS FOR MULTIPLE RELATION PATTERNS

Knowledge Graph Embedding (KGE) models that rely on translation operations, such as TransE, excel at capturing antisymmetric relations but fall short when it comes to modeling symmetric relations. On the other hand, tensor decomposition-based models like RESCAL, as previously discussed, can naturally represent symmetric relations due to the commutative property of matrix multiplication. However, these models struggle with antisymmetric relations and complex patterns, including inverse and composite relations. Current models capable of handling various relation patterns primarily focus on enhancing tensor decomposition models, relation-aware mapping models, and those that treat relations as rotations between entities.

### 4.1 Modified Tensor Decomposition-Based Models

To overcome the limitation that real space matrix multiplication cannot effectively model antisymmetric relations, modified tensor decomposition-based KGE models have been developed to handle both symmetric and antisymmetric relations. ComplEx [45] pioneers the embedding of entities and relations into complex space, enhancing DistMult by employing Hamiltonian multiplication among the complex representations of head entities, relations, and tail entities. The scoring function is defined as:

$$E(h,r,t) = Re(h^T diag(r)\bar{t}) \tag{30}$$

where $h \in \mathbb{C}^k$ and $t \in \mathbb{C}^k$ are the complex vector embeddings of the head and tail entities, respectively. $r \in \mathbb{C}^k$ is the complex vector embedding of the relation, $\bar{t}$ denotes the conjugate of $t$, and $Re$ represents taking the real part of a complex number. By calculating the scores of the triples $(h, r, t)$ and $(t, r, h)$ using Eq. (30), the conjugate of the complex vector of the tail entity $t$ is taken in the score of the triple $(h, r, t)$, while the conjugate of the complex vector of the head entity $h$ is taken in the score of the triple $(t, r, h)$. This allows these two triples to potentially yield the same or different scores, achieving the simultaneous modeling of both symmetric and antisymmetric relations.

HolE [46] diverges from models like RESCAL and DistMult, which use symmetric bilinear operations for the head and tail entities, by employing circular correlation operations between the vector embeddings of an entity pair. This makes the triple scoring asymmetric with respect to the head and tail entities. The scoring function of HolE is designed as follows:

$$E(h, r, t) = \boldsymbol{r}^T (\boldsymbol{h} \star \boldsymbol{t}) \tag{31}$$

$$[\boldsymbol{h} \star \boldsymbol{t}]_i = \sum_{j=0}^{k-1} [\boldsymbol{h}]_j [\boldsymbol{t}]_{i+j \bmod k} \tag{32}$$

where $\boldsymbol{h} \in \mathbb{R}^k$ and $\boldsymbol{t} \in \mathbb{R}^k$ are the vector embeddings of the head and tail entities, respectively. $\boldsymbol{r} \in \mathbb{R}^k$ is the vector embedding of the relation, and $\star$ denotes the circular correlation operator. Post the circular correlation operation on the head and tail entities, the i-th element is derived from the j-th element of the head entity vector and the (i+j)-th element of the tail entity vector. When the positions of the head and tail entities are swapped, the circular correlation operation results in the ii-th element being derived from the jj-th element of the tail entity vector and the (i+j)-th element of the head entity vector. Thus, the scores of the triples can be the same or different depending on the order of the entities, achieving a similar effect to ComplEx and enabling the simultaneous modeling of symmetric and antisymmetric relations.

SimplE [47] extends DistMult by learning embeddings for each entity as both head and tail entities and by constructing an inverse relation for each relation. For a triple (h,r,t)(h,r,t) and its corresponding triple with the inverse relation (t,r−1,h)(t,r−1,h), both should hold true. Therefore, the SimplE model scores both the original triple and the triple with the inverse relation. The scoring function is defined as follows:

$$E(h, r, t) = \boldsymbol{h}_h^T diag(\boldsymbol{r}) \boldsymbol{t}_t + \boldsymbol{t}_h^T diag(\boldsymbol{r}^{-1}) \boldsymbol{h}_t \tag{33}$$

where $\boldsymbol{h}_h \in \mathbb{R}^k$ and $\boldsymbol{h}_t \in \mathbb{R}^k$ represent the vector embeddings of the head entity $h$ as both a head and tail entity, respectively. Similarly, $\boldsymbol{t}_h \in \mathbb{R}^k$ and $\boldsymbol{t}_t \in \mathbb{R}^k$ represent the vector embeddings of the tail entity $t$ as both a head and tail entity. $diag(\boldsymbol{r}) \in \mathbb{R}^{k \times k}$ and $diag(\boldsymbol{r}^{-1}) \in \mathbb{R}^{k \times k}$ are the diagonal matrix representations of the relation $r$ and its inverse $r^{-1}$, respectively. Notably, when scoring triples with swapped head and tail entities, the bilinear operation involving the same relation uses different vector embeddings for the head and tail entities. This allows for the triples to have the same or different scores, effectively modeling both symmetric and antisymmetric relations.

For a visual aid, the key concepts of several improved tensor decomposition models, including ComplEx, HolE, and SimplE, are depicted in Figure 6.

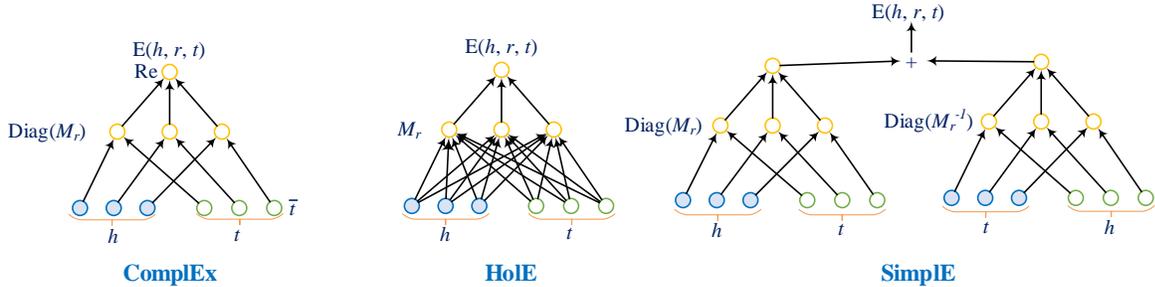

Figure 6: Illustrative diagram of the models based on modified tensor decomposition

These modified tensor decomposition models are capable of simultaneously modeling symmetric and antisymmetric relations. Specifically, for antisymmetric relations, the vector embeddings of the head and tail entities that undergo matrix multiplication with the same relation representation change when calculating triple scores by swapping the head and tail entities. This characteristic is also applicable to modeling inverse relation patterns between two relations. However, these approaches are limited in their ability to model patterns that involve the combination of two relations.

## 4.2 Models Based on Modified Relation-Aware Mapping

PairRE [48] addresses the joint modeling of relation patterns and the complex mapping characteristics of relations by utilizing paired relation embeddings. Its scoring function is defined as:

$$E(h, r, t) = \left\| \boldsymbol{h} \circ \boldsymbol{r}^h - \boldsymbol{t} \circ \boldsymbol{r}^t \right\| \tag{34}$$

where $\boldsymbol{h} \in \mathbb{R}^k$ and $\boldsymbol{t} \in \mathbb{R}^k$ are the vector embeddings of the head and tail entities, respectively. $\boldsymbol{r}^h \in \mathbb{R}^k$ and $\boldsymbol{r}^t \in \mathbb{R}^k$ are the paired vector embeddings of the relation, with $\|\boldsymbol{h}\|^2 = \|\boldsymbol{t}\|^2 = 1$. This approach allows the head and tail entity vectors to be mapped using different relation vectors. For instance, in a 1-N relation, PairRE can automatically adjust rtrt to consist of smaller element values, effectively reducing the weight related to the tail entity vector tt. This enables different tail entity embeddings corresponding to the same head entity and 1-N relation to meet the optimization target, achieving the goal of modeling complex relation mappings. Additionally, PairRE can model various relation patterns by ensuring the relation embeddings satisfy the following constraints: (1) Symmetric relations: $\boldsymbol{r}^{h2} = \boldsymbol{r}^{t2}$. (2) Antisymmetric relations: $\boldsymbol{r}^{h2} \neq \boldsymbol{r}^{t2}$. (3) Inverse relations: $\boldsymbol{r}_1^h \circ \boldsymbol{r}_2^h = \boldsymbol{r}_1^t \circ \boldsymbol{r}_2^t$. (4) Composite relations: $\boldsymbol{r}_1^h \circ \boldsymbol{r}_2^h \circ \boldsymbol{r}_3^h = \boldsymbol{r}_1^t \circ \boldsymbol{r}_2^t \circ \boldsymbol{r}_3^t$.

TripleRE [49] extends the PairRE model by representing relations as mappings and translations for both the head and tail entities. Its scoring function is designed as follows:

$$E(h, r, t) = \left\| \boldsymbol{h} \circ \boldsymbol{r}^h - \boldsymbol{t} \circ \boldsymbol{r}^t + \boldsymbol{r}^m \right\| \tag{35}$$

where $\boldsymbol{r}^m \in \mathbb{R}^k$ represents the relation vector for translating entities, while all other symbols are as defined in PairRE.

TranS [50] further extends TripleRE by requiring two mapping operations and three translation operations. The scoring function is defined as follows:

$$E = \left\| \boldsymbol{h} \circ \boldsymbol{t}^h - \boldsymbol{t} \circ \boldsymbol{h}^t + \boldsymbol{r}^h \circ \boldsymbol{h} + \boldsymbol{r} + \boldsymbol{r}^t \circ \boldsymbol{t} \right\| \tag{36}$$

in which $\boldsymbol{t}^h \in \mathbb{R}^k$ represents the auxiliary tail entity vector, $\boldsymbol{h}^t \in \mathbb{R}^k$ represents the auxiliary head entity vector, $\boldsymbol{r}^h \in \mathbb{R}^k$ represents the auxiliary relation vector associated with the head entity, and $\boldsymbol{r}^t \in \mathbb{R}^k$ represents the auxiliary relation vector associated with the tail entity.

## 4.3 Rotation-Based Models

Rotation-based KGE models aim to simultaneously capture symmetric, antisymmetric, inverse, and composite relation patterns by combining the strengths of translation-based models for antisymmetric, inverse, and composite relations with those of improved tensor decomposition models for symmetric relations.

RotatE [51] pioneered the approach of viewing the transformation from the head entity to the tail entity as a rotation operation. This model embeds entities and relations into the complex vector space and employs the Hadamard product to achieve this rotation. The scoring function of RotatE is designed as follows:

$$E(h, r, t) = \left\| \boldsymbol{h} \circ \boldsymbol{r} - \boldsymbol{t} \right\| \tag{37}$$

where $\boldsymbol{h} \in \mathbb{C}^k$ and $\boldsymbol{t} \in \mathbb{C}^k$ are the complex vector embeddings of the head and tail entities, respectively. $\boldsymbol{r} \in \mathbb{C}^k$ is the complex vector embedding of the relation, and $\circ$ denotes the Hadamard product, which calculates the element-wise product of two vectors. Specific to a triple $(h, r, t)$, the scoring function of RotatE is $\boldsymbol{h} \circ \boldsymbol{r} = \boldsymbol{t}$. Based on this optimization objective, RotatE has been theoretically proven to model various relation patterns, including symmetric, antisymmetric, inverse, and composite relations. The relation embeddings should satisfy specific constraints: (1) Symmetric relations: $\boldsymbol{r} \circ \boldsymbol{r} = \mathbf{1}$. (2) Antisymmetric relations: $\boldsymbol{r} \circ \boldsymbol{r} \neq \mathbf{1}$. (3) Inverse relations: $\boldsymbol{r}_1 = \boldsymbol{r}_2^{-1}$. (4) Composition relations: $\boldsymbol{r}_1 = \boldsymbol{r}_2 \circ \boldsymbol{r}_3$.

However, RotatE implements rotations on a complex plane using Euler angles in complex vector space, and encounters singularity issues. To ensure more stable rotation operations, QuatE [52] embeds entities and relations in quaternion space and achieves rotations on planes with Hamilton multiplication. The scoring function of QuatE is defined as follows:

$$E(h, r, t) = \boldsymbol{h} \otimes (\boldsymbol{r}/|\boldsymbol{r}|) \cdot \boldsymbol{t} \tag{38}$$

where $\boldsymbol{h} \in \mathbb{Q}^k$ and $\boldsymbol{t} \in \mathbb{Q}^k$ are the k-dimensional quaternion vectors of the head and tail entities, respectively. $\boldsymbol{r} \in \mathbb{Q}^k$ is the k-dimensional quaternion vector of the relation, and $|\boldsymbol{r}|$ represents the norm of $\boldsymbol{r}$. A quaternion consists of one real part and three imaginary parts. $\otimes$ denotes Hamilton multiplication, and $\cdot$ represents the dot product of complex vectors. Notably, while RotatE can model composite relation patterns involving three different relations, it struggles with patterns like $Brother(x, z) \Leftarrow Brother(x, y) \wedge Sister(y, z)$ that involve two of the same relations. In contrast, QuatE can model composite relation patterns involving two of the same relations.

DualE [53] enhances the modeling capabilities of interactions between entities and relations by combining rotation and translation operations. It represents entities and relations in dual quaternion space and uses dual quaternion multiplication to express rotations and translations. The scoring function is defined as follows:

$$E(h, r, t) = \langle \boldsymbol{Q}_h \underline{\otimes} \boldsymbol{W}_h, \boldsymbol{Q}_t \rangle \tag{39}$$

where $\boldsymbol{Q}_h = \boldsymbol{a} + \epsilon\boldsymbol{b}$, $\boldsymbol{W}_h = \boldsymbol{c} + \epsilon\boldsymbol{d}$, and $\boldsymbol{Q}_t = \boldsymbol{e} + \epsilon\boldsymbol{f}$ represent the dual quaternion vector embeddings of the head entity, relation, and tail entity, respectively. $\boldsymbol{a}$, $\boldsymbol{b}$, $\boldsymbol{c}$, $\boldsymbol{d}$, $\boldsymbol{e}$ and $\boldsymbol{f}$ are quaternion vectors. $\epsilon$ is the dual unit that satisfies $\epsilon^2 = 0$. $\underline{\otimes}$ denotes dual quaternion multiplication, and $\langle \boldsymbol{x}, \boldsymbol{y} \rangle$ represents the dot product of dual quaternions $\boldsymbol{x}$ and $\boldsymbol{y}$.

To achieve combined rotation and translation effects similar to DualE, BiQUE [54] uses dual quaternions to represent each relation. One quaternion is used to linearly add with the head entity embedding for translation, and the other quaternion is used to rotate the translated embedding using Hamilton multiplication. The scoring function for BiQUE is defined as:

$$E(h, r, t) = \langle (\boldsymbol{Q}_h + \boldsymbol{Q}_r^+) \otimes \boldsymbol{Q}_r^\times, \boldsymbol{Q}_t \rangle \tag{40}$$

where $\boldsymbol{Q}_h \in \mathbb{Q}^k$ and $\boldsymbol{Q}_t \in \mathbb{Q}^k$ represent the quaternion vector embeddings of the head and tail entities, respectively. $\boldsymbol{Q}_r^+ \in \mathbb{Q}^k$ and $\boldsymbol{Q}_r^\times \in \mathbb{Q}^k$ are the two quaternion representations of the relation $r$ used for translation and rotation operations.

DihEdral [55] utilizes the dihedral group to represent relations, constructing its elements through rotation and reflection operations on a 2D symmetric polygon. The number of rotation elements $KK$ determines the minimum angle of rotation around the center, $2\pi KK2\pi$, thus encapsulating the characteristic of discrete rotation operations. The scoring function of this model is defined as follows:

$$E(h, r, t) = \sum_{l=0}^{L} \boldsymbol{h}^{(l)T} \boldsymbol{R}^{(l)} \boldsymbol{t}^{(l)} \tag{41}$$

where $\boldsymbol{h} \in \mathbb{R}^{2L}$ and $\boldsymbol{t} \in \mathbb{R}^{2L}$ are the vector embeddings of the head and tail entities. $\boldsymbol{h}^{(l)} \in \mathbb{R}^2$ and $\boldsymbol{t}^{(l)} \in \mathbb{R}^2$ are the $l$-th components of the head and tail entity vectors, respectively. $\boldsymbol{R}^{(l)} \in \mathrm{D}_k$ is the dihedral group matrix representation of the $l$-th component of the relation.

CompoundE [56] extends DualE by establishing translation, rotation, and scaling operations simultaneously in dual quaternion space. Its scoring function is defined as follows:

$$E(h, r, t) = \|\boldsymbol{T}_r \cdot \boldsymbol{R}_r \cdot \boldsymbol{S}_r \cdot \boldsymbol{h} - \boldsymbol{t}\| \tag{42}$$

where $\boldsymbol{T}_r$, $\boldsymbol{R}_r$, and $\boldsymbol{S}_r$ represent the translation, rotation, and scaling matrices corresponding to the relation $r$.

HA-RotatE [57] is developed upon RotatE by defining a weight parameter for each relation. This strategy models various relation patterns and simultaneously leverages the weight parameter to control the modeling of hierarchical relations between entities. The scoring function of HA-RotatE is defined as:

$$E(h, r, t) = \|\boldsymbol{W}_r \boldsymbol{h} \circ \boldsymbol{r} - \boldsymbol{t}\| \tag{43}$$

where $\boldsymbol{W}_r$ is the weight parameter for relation $r$.

Given that both modified relation-aware mapping models and rotation-based models can be perceived as moving entities within the representation space, Figure 7 illustrates the characteristics of several typical models.

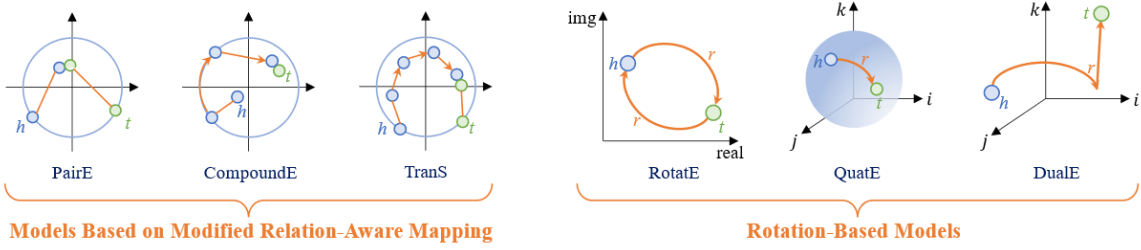

Figure 7: Illustrative diagram of the models that could model relation patterns

Furthermore, Table 2 provides a comprehensive summary of the three types of models discussed in this section for modeling various relation patterns.

Table 2: A comprehensive summary of models modeling various relation patterns

| Model | | Embedding Space | Scoring Function | Characteristics | Pros and Cons |
|---|---|---|---|---|---|
| Modified Tensor Decomposition-Based Models | ComplEx | Complex Vector Space | $Re(\boldsymbol{h}^T diag(\boldsymbol{r})\bar{\boldsymbol{t}})$ | Using conjugate of complex numbers in triple scoring to model both symmetric and antisymmetric relations. | Pros: allowing for the clever modeling of both symmetric and antisymmetric relations.  Cons: cannot model the combination patterns of two relations. |
| | HolE | Real Vector Space | $\boldsymbol{r}^T(\boldsymbol{h} \star \boldsymbol{t})$ | Circular correlation operations between head and tail entity vectors to model symmetric and antisymmetric relations. | |
| | SimplE | | $\boldsymbol{h}_h^T diag(\boldsymbol{r})\boldsymbol{t}_t$ $+ \boldsymbol{t}_h^T diag(\boldsymbol{r}^{-1})\boldsymbol{h}_t$ | Learning embeddings for each entity as both head and tail entities and constructing an inverse relation for each relation. | |
| Models Based on Modified Relation-Aware Mapping | PairRE | Real Vector Space | $\|\boldsymbol{h} \circ \boldsymbol{r}^h - \boldsymbol{t} \circ \boldsymbol{r}^t\|$ | Using paired relation vector embeddings to simultaneously model relation patterns and complex mappings. | Pros: model various relation patterns, but  Cons: require additional mapping and translation operations, leading to a larger number of parameters. |
| | TripleRE | | $\|\boldsymbol{h} \circ \boldsymbol{r}^h - \boldsymbol{t} \circ \boldsymbol{r}^t + \boldsymbol{r}^m\|$ | Representing relations as mappings and translations for both head and tail entities. | |
| | TranS | | $\|\boldsymbol{h} \circ \boldsymbol{t}^h - \boldsymbol{t} \circ \boldsymbol{h}^t + \boldsymbol{r}^h \circ \boldsymbol{h} + \boldsymbol{r} + \boldsymbol{r}^t \circ \boldsymbol{t}\|$ | Requiring two mappings and three translation operations for head and tail entities. | |
| Rotation-Based Models | RotatE | Complex Vector Space | $\|\boldsymbol{h} \circ \boldsymbol{r} - \boldsymbol{t}\|$ | First to view the transformation from head to tail entity as a rotation operation on a plane. | Pros: allowing for the clever modeling of both symmetric and antisymmetric relations.  Cons: cannot model the combination patterns of two relations. |
| | QuatE | Quaternion Space | $\boldsymbol{h} \otimes (\boldsymbol{r}/|\boldsymbol{r}|) \cdot \boldsymbol{t}$ | Representing entities and relations in quaternion space for rotations on two planes. | |
| | DualE | Dual Quaternion Space | $\langle \boldsymbol{Q}_h \underline{\otimes} \boldsymbol{W}_h, \boldsymbol{Q}_t \rangle$ | Using dual quaternion space to represent entities and relations with rotation and translation operations. | |

| BiQUE | Quaternion Space | $\langle (\boldsymbol{Q}_h + \boldsymbol{Q}_r^+) \otimes \boldsymbol{Q}_r^\times, \boldsymbol{Q}_t \rangle$ | Using two quaternions to represent each relation for rotation and translation. | |
|---|---|---|---|---|
| DihEdral | Dihedral Group | $\sum_{l=0}^{L} \boldsymbol{h}^{(l)T} \boldsymbol{R}^{(l)} \boldsymbol{t}^{(l)}$ | Establishing discrete rotation and reflection operations in the dihedral group. | |
| CompoundE | Dual Quaternion Space | $\|\boldsymbol{T}_r \cdot \boldsymbol{R}_r \cdot \boldsymbol{S}_r \cdot \boldsymbol{h} - \boldsymbol{t}\|$ | Simultaneously establishing translation, rotation, and scaling operations. | |
| HA-RotatE | Complex Vector Space | $\|\boldsymbol{W}_r \boldsymbol{h} \circ \boldsymbol{r} - \boldsymbol{t}\|$ | Applying linear mapping before rotation to model various relation patterns and hierarchical relations. | |

## 5 MODELS FOR HIERARCHICAL RELATIONS

Capturing the inherent hierarchical relations between entities in KGs is a critical aspect of Knowledge Graph Embedding (KGE). Current research primarily focuses on three approaches: leveraging additional auxiliary information, employing hyperbolic space, and utilizing polar coordinate systems.

### 5.1 Auxiliary Information-Based Models

HCE [58] capitalizes on the hierarchical structure of entity types within the ontology layer of KGs. It concurrently learns entity and type embeddings, predicated on the task of predicting contextual entities. HCE introduces the ancestor categories of an entity's type when learning an entity's embedding. If a type embedding is in proximity to an entity embedding, the ancestor type embeddings of that type will also be close to the entity embedding, thereby integrating hierarchical relations between types into the entity embeddings.

HRS [59] concentrates on the hierarchy among relations. It clusters relations into higher-level relations and extends TransE to calculate the difference between the tail and head entity vectors in the triples associated with each relation. These differences are then clustered to derive lower-level sub-relations, with the original relations forming the middle layer. This method jointly represents the translation operation from the head entity to the tail entity across three levels of relations. The scoring function is defined as follows:

$$E(h, r, t) = \|\boldsymbol{h} + \boldsymbol{r}_c + \boldsymbol{r} + \boldsymbol{r}_s - \boldsymbol{t}\|_{L1/L2} \tag{44}$$

where $\boldsymbol{r}_c$, $\boldsymbol{r}$, and $\boldsymbol{r}_s$ represent the vector embeddings of higher-level clustered relations, middle-layer relations, and lower-level sub-relations, respectively.

TKRL [60] learns the embeddings of all entity types at different levels through hierarchical type mapping matrices, thereby transforming a triple into associations among multiple hierarchical entity types. Its scoring function is defined as follows:

$$E(h, r, t) = \|\boldsymbol{M}_h \boldsymbol{h} + \boldsymbol{r} - \boldsymbol{M}_t \boldsymbol{t}\|_{L1/L2} \tag{45}$$

where $\boldsymbol{M}_h$ and $\boldsymbol{M}_t$ are the hierarchical type mapping matrices. TKRL designs two hierarchical type encoding mechanisms. The first is cyclic hierarchical encoding, which maps an entity into the most specific lower-level entity type representation space first, then into higher-level entity type representation spaces. This mechanism gradually and cyclically maps an entity from lower-level types to higher-level types. The type mapping matrix for cyclic hierarchical encoding is defined as follows:

$$\boldsymbol{M}_c = \boldsymbol{M}_{c^{(1)}} \boldsymbol{M}_{c^{(2)}} \cdots \boldsymbol{M}_{c^{(n)}} \tag{46}$$

in which $n$ represents the level of an entity type within the overall hierarchical structure, and $\boldsymbol{M}_{c^{(i)}}$ denotes the type mapping matrix of the i-th level, with the first level being the lowest. The second hierarchical type encoding mechanism is weighted hierarchical encoding, where different weights reflect the hierarchical structure of types. The type mapping matrix for weighted hierarchical encoding is defined as follows:

$$\boldsymbol{M}_c = \beta_1 \boldsymbol{M}_{c^{(1)}} + \beta_2 \boldsymbol{M}_{c^{(2)}} + \cdots + \beta_n \boldsymbol{M}_{c^{(n)}} \tag{47}$$

where $\beta_i$ represents the weight of the ii-th level type. To enhance the hierarchy among types, considering that higher-level types convey generalized abstract semantics while lower-level types express precise semantics, a strategy is designed where weights decrease from lower-level types to higher-level types:

$$\beta_i : \beta_{i+1} = (1 - \eta) : \eta, \qquad \sum_{i=1}^{n} \beta_i = 1 \tag{48}$$

in which the parameter $\eta$ is set within the range $\eta \in (0, 0.5)$.

### 5.2 Hyperbolic Space-Based Models for Hierarchical Relations

The hierarchical structure inherent in KGs can be effectively modeled using hyperbolic space, which offers a more natural representation for tree-like hierarchies than Euclidean space. Hyperbolic embeddings enable high-level, abstract entities to be represented with lower-dimensional vectors, while more specific, lower-level entities are captured with higher-dimensional vectors.

Poincare Embeddings [61] pioneered the embedding of KGs into hyperbolic space, particularly focusing on "is-a" hierarchical relations. By mapping the KG into the Poincare ball, a distinctive geometric space, this model places higher-level entities closer to the center of the ball and lower-level entities further away. The scoring function for Poincare embeddings is given by:

$$E(h, t) = arcosh(1 + 2 \frac{\|\boldsymbol{h} - \boldsymbol{t}\|^2}{(1 - \|\boldsymbol{h}\|^2)(1 - \|\boldsymbol{t}\|^2)}) \tag{49}$$

where $\boldsymbol{h} \in \mathcal{B}^k$ and $\boldsymbol{t} \in \mathcal{B}^k$ are the vector embeddings of the head and tail entities in the Poincare ball space, respectively.

MuRP [62] extends Poincare to accommodate more complex hierarchical structures with a variety of relations. It combines the geometry of the Poincare ball with a triple scoring function, resulting in a scoring function that can handle multiple relations:

$$E(h, r, t) = -d_B\left(\boldsymbol{h}^{(r)}, \boldsymbol{t}^{(r)}\right)^2 + b_h + b_t = -d_B(\exp_0^c(\boldsymbol{R}\log_0^c(\boldsymbol{h})), \boldsymbol{t} \oplus_c \boldsymbol{r})^2 + b_h + b_t \tag{50}$$

in which $\boldsymbol{h} \in \mathcal{B}^k$ and $\boldsymbol{t} \in \mathcal{B}^k$ are the hyperbolic embeddings of the head and tail entities in the Poincare ball space. $\boldsymbol{R} \in \mathcal{B}^{k \times k}$ is the diagonal matrix corresponding to relation $r$, and $\boldsymbol{r} \in \mathcal{B}^k$ represents the hyperbolic translation vector for relation $r$. The hyperbolic embedding of the head entity $\boldsymbol{h}^{(r)} \in \mathcal{B}^k$ is obtained through Mobius matrix-vector multiplication to interact with the relation, while the hyperbolic embedding of the tail entity $\boldsymbol{t}^{(r)} \in \mathcal{B}^k$ is obtained through Mobius addition $\oplus_c$ to interact with the relation.

Since MuRP represents entities and relations in a hyperbolic space with fixed curvature, it cannot model multiple relation patterns. ATTH [63] leverages the more expressive rotation and reflection operations from DihEdral and incorporates an attention mechanism to adjust the importance of rotation and reflection operations. It simultaneously models multiple relation patterns and hierarchical relations between entities. The scoring function of ATTH is defined as follows:

$$E(h, r, t) = -d_B\left(\boldsymbol{h}^{(r)}, \boldsymbol{t}\right)^2 + b_h + b_t \tag{51}$$

$$\boldsymbol{h}^{(r)} = Att(\boldsymbol{h}_{Rot}, \boldsymbol{h}_{Ref}) \oplus_c \boldsymbol{r} \tag{52}$$

$$Att(\boldsymbol{h}_{Rot}, \boldsymbol{h}_{Ref}) = \exp_0^c(\boldsymbol{\alpha}_r^1 \boldsymbol{h}_{Rot} + \boldsymbol{\alpha}_r^2 \boldsymbol{h}_{Ref}) \tag{53}$$

$$(\boldsymbol{\alpha}_r^1, \boldsymbol{\alpha}_r^2) = \text{softmax}(\boldsymbol{\alpha}^T \log_0^c(\boldsymbol{h}), \boldsymbol{\alpha}^T \log_0^c(\boldsymbol{t}))$$

where $\boldsymbol{h}^{(r)} \in \mathcal{B}^k$ represents the hyperbolic embedding of the head entity obtained through weighted rotation and reflection operations using an attention mechanism. $\boldsymbol{h}_{Rot} \in \mathcal{B}^k$ is the head entity embedding after the relation-associated rotation operation, and $\boldsymbol{h}_{Ref} \in \mathcal{B}^k$ is the head entity embedding after the relation-associated reflection operation. $\boldsymbol{\alpha}_r^1 \in \mathcal{B}^k$ and $\boldsymbol{\alpha}_r^2 \in \mathcal{B}^k$ are the attention weight parameters associated with $\boldsymbol{h}_{Rot}$ and $\boldsymbol{h}_{Ref}$, respectively. $Att$ represents the attention mechanism.

Similar to the foundational role of TransE in the series of translation-based models, the series of hyperbolic space-based models are various extensions of MuRP. Among them, HyperKA [64] uses Mobius addition to represent the translation operation of hyperbolic embeddings of entities, followed by the aggregation of entity neighborhood information using

Graph Neural Networks (GNN) [65]. MuRMP [66] combines mixed curvature models with GNN to address the issue of representing entities and relations in a hyperbolic space with fixed curvature in MuRP. UltraE [67] constructs a hyperbolic manifold space by combining hyperbolic and manifold spaces to simultaneously represent relations at different levels and the same level.

### 5.3 Polar Coordinates-Based Models

HAKE [68] is the pioneering model that employs polar coordinates for KGE. In the polar coordinate system, a point in 2D space is defined by its magnitude and phase angle. HAKE transforms the head entity to the tail entity through operations specific to each relation, where the magnitude component represents scaling at a fixed angle, and the phase angle component signifies rotation at a fixed magnitude. This allows the magnitude to model entities at different hierarchical levels, while the phase angle models different entities at the same level. The scoring function of HAKE is defined as follows:

$$E(h, r, t) = d_{r,m}(\boldsymbol{h}_m, \boldsymbol{t}_m) + \lambda d_{r,p}(\boldsymbol{h}_p, \boldsymbol{t}_p) \tag{54}$$

$$d_{r,m}(\boldsymbol{h}_m, \boldsymbol{t}_m) = \|\boldsymbol{h}_m \circ \boldsymbol{r}_m - \boldsymbol{t}_m\|_{L2} \tag{55}$$

$$d_{r,p}(\boldsymbol{h}_p, \boldsymbol{t}_p) = \left\| \sin \left( (\boldsymbol{h}_p + \boldsymbol{r}_p - \boldsymbol{t}_p)/2 \right) \right\|_{L1} \tag{56}$$

where $d_{r,m}(\boldsymbol{h}_m, \boldsymbol{t}_m)$ and $d_{r,p}(\boldsymbol{h}_p, \boldsymbol{t}_p)$ represent the scores of the triple based on magnitude and phase angle, respectively, with $\lambda$ being the weighting parameter that balances these two scores. $\boldsymbol{h}_m \in \mathbb{R}^k$ and $\boldsymbol{t}_m \in \mathbb{R}^k$ are the magnitude vector embeddings of the head and tail entities, respectively. $\boldsymbol{h}$ and $\boldsymbol{t} \in \mathbb{R}^k$, respectively. $\boldsymbol{r}_m \in \mathbb{R}_+^k$ is the magnitude vector embedding with all positive elements. $\boldsymbol{h} \in [0, 2\pi)^k$, $\boldsymbol{r} \in [0, 2\pi)^k$, and $\boldsymbol{t} \in [0, 2\pi)^k$ are the phase angle vector embeddings of the head entity, relation, and tail entity, respectively.

Notably, HAKE's core idea is that the higher the semantic level of an entity, the shorter its magnitude; conversely, the lower the semantic level, the longer its magnitude. During model training, the magnitude representation of relations tends to conform to the following constraints: (1) The head entity has a higher semantic level than the tail entity: $[\boldsymbol{r}_m]_i > 1$. (2) The head and tail entities are at the same semantic level: $[\boldsymbol{r}_m]_i = 1$. (3) The head entity has a lower semantic level than the tail entity: $[\boldsymbol{r}_m]_i < 1$.

H2E [69] effectively combines hyperbolic space and polar coordinates for modeling hierarchical relations by learning hyperbolic polar coordinate embeddings of entities in hyperbolic space, utilizing the concepts of magnitude and phase angle. Similarly, HBE [70] employs Mobius multiplication and Mobius addition in an extended Poincare ball space to achieve rotation concerning phase angle and scaling concerning magnitude in polar coordinates, thereby modeling hierarchical relations between entities.

Several models capable of modeling hierarchical relations between entities are illustrated in Figure 8. Auxiliary information-based TKRL enables entities to map different levels of types starting from lower-level types. Hyperbolic space-based model MuRP allows for mapping and translation operations in Poincare ball space. Polar coordinates-based HAKE represents relations as magnitude $\boldsymbol{r}_m$ and phase angle $\boldsymbol{r}_p$, transforming the head entity to the tail entity by altering magnitude and phase angle.

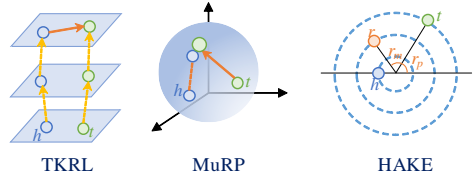

Figure 8: Illustrative diagram of the models representing hierarchical relations between entities

Besides, Table 3 provides a summary and comparison of various models that can model hierarchical relations between entities, particularly focusing on the characteristics, advantages, and disadvantages of models based on auxiliary information, hyperbolic space, and polar coordinates.

Table 3: A comprehensive summary of models modeling various relation patterns

| Model | | Embedding Space | Scoring Function | Characteristics | Pros and Cons |
|---|---|---|---|---|---|
| Auxiliary Information-Based Models | HRS | Real Vector Space | $\|\boldsymbol{h} + \boldsymbol{r}_c + \boldsymbol{r} + \boldsymbol{r}_s - \boldsymbol{t}\|_{L1/L2}$ | Using three layers of relations to represent the translation operation from the head entity to the tail entity. | Pros: directly represents hierarchical relations between entities.

Cons: completely relies on the auxiliary information, making it unsuitable for KGs lacking such information. |
| | TKRL | | $\|\boldsymbol{M}_r\boldsymbol{h} + \boldsymbol{r} - \boldsymbol{M}_t\boldsymbol{t}\|_{L1/L2}$ $\boldsymbol{M}_c = \boldsymbol{M}_{c^{(1)}}\boldsymbol{M}_{c^{(2)}} \cdots \boldsymbol{M}_{c^{(n)}}$ | Learning representations of all entity types at different levels using hierarchical type mapping matrices. | |
| Hyperbolic Space-Based Models | Poincare | Hyperbolic Space | $arcosh(1 + 2\dfrac{\|\boldsymbol{h} - \boldsymbol{t}\|^2}{(1 - \|\boldsymbol{h}\|^2)(1 - \|\boldsymbol{t}\|^2)})$ | Using the Poincare ball to represent entities; embeddings closer to the center represent higher-level entities, but only for is-a hierarchical relation. | Pros: representing hierarchical relations through the natural continuous tree structure of hyperbolic space.

Cons: the hierarchical characteristics are less significant compared to the other two categories. |
| | MuRP | | $\|\boldsymbol{h} \circ \boldsymbol{r}^h - \boldsymbol{t} \circ \boldsymbol{r}^t + \boldsymbol{r}^m\|$ | Combining triple scoring functions with the Poincare ball geometric to model different hierarchical structures of relations. | |
| | ATTH | | $-d_B(\boldsymbol{h}^{(r)}, \boldsymbol{t})^2 + b_h + b_t$ | Modeling rotation and reflection operations in hyperbolic space, modeling both multiple relation patterns and hierarchical relations between entities. | |
| Polar Coordinates-Based Models | HAKE | Complex Vector Space | $d_{r,m}(\boldsymbol{h}_m, \boldsymbol{t}_m) + \lambda d_{r,p}(\boldsymbol{h}_p, \boldsymbol{t}_p)$ | In polar coordinates, magnitude can model entities at different levels, and phase angle is able to model entities at the same level. | Pros: magnitude in polar coordinates can represent hierarchical characteristics.

Cons: it weakens the ability to model relation patterns. |
| | H2E | Quaternion Space | $d_B(\boldsymbol{h}_m, \boldsymbol{t}_m) + \lambda d_B(\boldsymbol{h}_p, \boldsymbol{t}_p)$ | Using the concepts of magnitude and phase angle in hyperbolic space to learn hyperbolic polar coordinate | |

| | | | embeddings of entities. | |
|---|---|---|---|---|
| HBE | Dual Quaternion Space | $\alpha\|2\tanh^{-1}((\boldsymbol{R} \otimes_c \boldsymbol{h})\oplus_c - (\boldsymbol{t}\oplus_c \boldsymbol{r}))\|^2 + \beta\|(\theta_h + \theta_r - \theta_t)\bmod 2\pi\|$ | Using Mobius multiplication and Mobius addition in an extended Poincare ball space to achieve rotation concerning phase angle and scaling concerning magnitude in polar coordinates. | |

## 6  FUTURE RESEARCH DIRECTIONS

Current research in KGE is heavily focused on modeling complex relation mappings, various relation patterns, and hierarchical relations between entities. While significant strides have been made, there is room for further advancement, particularly in enhancing the effectiveness of KGE models and their performance in downstream tasks such as knowledge graph completion. This section explores several promising avenues for future research, including the integration of multimodal information, the use of logical rules, and the consideration of dynamic KGs.

### 6.1  KGE with Multimodal Information

Existing KGE models primarily concentrate on embedding space, scoring functions, and encoding strategies for entities and relations. These models excel at capturing structural features for dense KGs but often struggle with sparse KGs or entities and relations with long-tail distributions due to limited graph structure data. Integrating multimodal information corresponding to entities and relations can significantly aid in establishing complex relation mappings and hierarchical relations. For example, in a 1-N relation involving multiple tail entities, textual descriptions may vary, but the relation's textual content remains similar. Extracting similar feature representations from textual descriptions can effectively model complex relation mappings. Similarly, visual features can be extracted from images of entities that exhibit hierarchical characteristics, such as *car* and *tire* in the triple (*car*, *part*, *tire*).

Models like DKRL [71] and KG-BERT [72] utilize textual descriptions, while IKRL [73] and RSME [74] extract visual features to enhance entity representations. These approaches improve the accuracy of entity and relation representations and the overall performance of KGE, especially in sparse KGs. Future research could explore the impact of these multimodal integration strategies on modeling relation characteristics.

### 6.2  Rule-Enhanced Relation Pattern Modeling

Current KGE models predominantly rely on automatically learning implicit relation patterns from triples. However, in sparse KGs, it is challenging to ensure that relation embeddings meet the constraints corresponding to each relation pattern. Most KGs possess an ontology that defines axioms or rules, which can be mined using rule learning algorithms to express logical associations between relations and reflect relation patterns directly. Models such as RPJE [75] and EngineKG [76] can mine rules from KGs that express inverse and composite relation patterns, enhancing the modeling capability for these patterns by injecting logical information into entity and relation embeddings. Future research could investigate the automatic mining of rules that express a broader range of relation patterns and utilize these rules to enhance entity and relation embeddings, thereby improving the modeling capability of relation patterns in sparse KGs.

### 6.3  Modeling Relation Characteristics for Dynamic KGs

Most current research on KGE for modeling relation characteristics focuses on static KGs. However, real-world data is often dynamic, and there is a paucity of research on modeling relation characteristics for dynamic KGs. Dynamic KGs frequently add new entities or relations with sparsely associated triples, making it difficult to learn various relation characteristics directly from limited data. Models like GANA [77] combine TransH's approach with meta-learning mechanisms to address complex relation modeling for few-shot KGE. The semantic information associated with entities and the non-fixed hierarchical relations between entities complicate modeling in dynamic KGs. While models such as DyERNIE [78] and HERCULES [79] improve upon MuRP and ATTH by embedding dynamic KGs into hyperbolic space for modeling hierarchical relations, their ability to model relation characteristics remains limited due to the lack of representation of temporal information.

Future research in this area could focus on developing models that not only capture the dynamic nature of KGs but also effectively incorporate temporal information to enhance the relation characteristics in evolving knowledge graphs.

## 7 CONCLUSION

Relations within KGs embody rich semantic information, and the accurate modeling of these relation characteristics is pivotal to the efficacy of KGE models. This paper commences with an introduction to the foundational concepts of KGE, delving into the nuances of relation characteristics, which include complex relation mappings, diverse relation patterns, and hierarchical relations among entities. Subsequently, we undertake a comprehensive review of models that leverage relation-aware mapping, specific representation spaces, tensor decomposition, and neural networks. These models are designed to adeptly capture complex relation mappings. Besides, we summarize models that employ modified tensor decomposition, modified relation-aware mapping, and rotation operations, which are tailored to model a variety of relation patterns. Furthermore, we elucidate models based on auxiliary information, hyperbolic space, and polar coordinates. These models are particularly effective in modeling hierarchical relations among entities, showcasing the diversity of approaches within the KGE landscape. In conclusion, this paper explores and discusses several promising future research directions for KGE, focusing on the modeling of relation characteristics. As the field continues to evolve, it is imperative for researchers to consider the dynamic and multifaceted nature of relations within KGs, ensuring that KGE models remain at the forefront of knowledge representation and reasoning.

## REFERENCES

[1] Z. Y. Liu, M. S. Sun, Y. K. Lin, et al. Research progress on knowledge representation learning. Journal of Computer Research and Development, vol. 53, no. 2, pp. 247-261, 2016.

[2] T. C. Zhang, X. Tian, X. H. Sun, et al. A comprehensive review of knowledge graph embedding techniques. Journal of Software, vol. 34, no. 01, pp. 277-311, 2023.

[3] X. Ge, Y. C. Wang, B. Wang, et al. Knowledge graph embedding: an overview. arXiv preprint arXiv:2309.12501, 2023.

[4] N. Zhang, S. Deng, Z. Sun, et al. Long-tail relation extraction via knowledge graph embeddings and graph convolution networks. Proceedings of the 17th Annual Conference of the North American Chapter of the Association for Computational Linguistics, pp. 3016-3025, 2019.

[5] Z. Wang, J. Yang, X. Ye. Knowledge graph alignment with entity-pair embedding. In Proceedings of the 2020 Conference on Empirical Methods in Natural Language Processing, pp. 1672-1680, 2020.

[6] G. Li, Z. Sun, W. Hu, et al. Position-aware relational transformer for knowledge graph embedding. IEEE Transactions on Neural Networks and Learning Systems, 2023.

[7] B. Y. Lin, X. Chen, J. Chen, et al. KagNet: Knowledge aware graph networks for commonsense reasoning. Proceedings of the 2019 Conference on Empirical Methods in Natural Language Processing and the 9th International Joint Conference on Natural Language Processing (EMNLP-IJCNLP), pp. 2829-2839, 2019.

[8] Y. Wang, A. Li, J. Zhang, et al. Enhanced knowledge graph embedding for multi-task recommendation via integrating attribute information and high-order connectivity. Proceedings of the 10th International Joint Conference on Knowledge Graphs, pp. 140-144, 2022.

[9] S. Pan, L. Luo. Unifying large language models and knowledge graphs: A roadmap. arXiv preprint arXiv:2306.08302, 2023.

[10] Q. Yan, J. Fan, M. Li, et al. A survey on knowledge graph embedding. Proceedings of the 7th IEEE International Conference on Data Science in Cyberspace (DSC), pp. 576-583, 2022.

[11] J. Cao, J. Fang, Z. Meng, et al. Knowledge graph embedding: A survey from the perspective of representation spaces. arXiv preprint arXiv:2211.03536, 2022.

[12] Q. Shen, H. Zhang, Y. Xu, H. Wang, K. Cheng. Comprehensive survey of loss functions in knowledge graph embedding models. Computer Science, vol. 50, no. 4, pp. 149-158, 2023.

[13] Z. Li, Y. Zhao, Y. Zhang. Survey of knowledge graph reasoning based on representation learning. Computer Science, vol. 50, no. 3, pp. 94-113, 2023.

[14] M. B. Yu, J. Q. Du, J. G. Luo, et al. Research progress of knowledge graph completion based on knowledge representation learning. Computer Engineering and Applications, vol. 59, no. 18, pp. 59-73, 2023.

[15]    X. Du, M. Liu, L. Shen, X. Peng. A survey of knowledge graph representation learning methods for link prediction. Journal of Software, vol. 35, no. 1, pp. 87-117, 2024.

[16]    D. Q. Nguyen. A survey of embedding models of entities and relationships for knowledge graph completion. Proceedings of the Graph-based Methods for Natural Language Processing (Text-Graphs), pp. 1-14, 2020.

[17]    Q. Wang, Z. Mao, B. Wang, et al. Knowledge graph embedding: A survey of approaches and applications. IEEE Transactions on Knowledge and Data Engineering, vol. 29, no. 12, pp. 2724-2343, 2017.

[18]    Y. Dai, S. Wang, N. N. Xiong, et al. A survey on knowledge graph embedding: Approaches, applications, and benchmarks. Electronics, vol. 9, no. 5, 2020.

[19]    G. A. Gesese, R. Biswas, H. Sack. A comprehensive survey of knowledge graph embeddings with literals: Techniques and applications. Workshop at ESWC 2020 on Deep Learning for Knowledge Graph, 2019.

[20]    M. Wang, L. Qiu, X. Wang. A survey on knowledge graph embeddings for link prediction. Symmetry, vol. 13, no. 3, 2021.

[21]    H. Wang, G. Qi, H. Chen. Knowledge graph methods, practices, and applications. Electronic Industry Press, 2019. (in Chinese)

[22]    S. Ji, S. Pan, E. Cambria, et al. A survey on knowledge graphs: Representation, acquisition, and applications. IEEE Transactions on Neural Networks and Learning Systems, vol. 33, no. 2, pp. 494-514.

[23]    M. M. Alam, R. R. M., M. Nayyeri, et al. Language model guided knowledge graph embeddings, in IEEE Access, vol. 10, pp. 76008-76020, 2022.

[24]    A. Bordes, N. Usunier, D. Garcia, et al. Translating embeddings for modeling multi-relational data. Proceedings of the 26th International Conference on Neural Information Processing Systems, pp. 2787-2795, 2013.

[25]    Z. Wang, J. Zhang, J. Feng, et al. Knowledge graph embedding by translating on hyperplanes. Proceedings of the 28th AAAI Conference on Artificial Intelligence, pp. 1112-1119, 2014.

[26]    Y. Lin, Z. Liu, M. Sun, et al. Learning entity and relation embeddings for knowledge graph completion. Proceedings of the 29th AAAI Conference on Artificial Intelligence, pp. 2181-2187, 2015.

[27]    D. Q. Nguyen, K. Sirts, L. Z. Qu, et al. STransE: A novel embedding model of entities and relationships in knowledge bases. Proceedings of the 2016 Conference of the North American Chapter of the Association for Computational Linguistics (NAACL), pp. 460-466.

[28]    G. L. Ji, S. Z. He, L. H. Xu, et al. Knowledge graph embedding via dynamic mapping matrix. Proceedings of the 53rd Annual Meeting of the Association for Computational Linguistics and the 7th International Joint Conference on Natural Language Processing. Beijing: Association for Computational Linguistics (ACL), pp. 687-696, 2015.

[29]    G. L. Ji, K. Liu, S. Z. He, et al. Knowledge graph completion with adaptive sparse transfer matrix. Proceedings of the 30th AAAI Conference on Artificial Intelligence (AAAI), pp. 985-991, 2016.

[30]    J. Feng, M. L. Huang, M. D. Wang, et al. Knowledge graph embedding by flexible translation. Proceedings of the 15th International Conference on Principles of Knowledge Representation and Reasoning (KR), pp. 557-560, 2016.

[31]    H. Xiao, M. L. Huang, Y. Hao, et al. TransA: An adaptive approach for knowledge graph embedding. arXiv:1509.05490, 2015.

[32]    M. Fan, Q. Zhou, E. Chang, et al. Transition-based knowledge graph embedding with relational mapping properties. Proceedings of the 28th Pacific Asia Conference on Language, Information and Computing (PACLIC), pp. 328-337, 2014.

[33]    S. He, K. Liu, G. Ji, et al. Learning to represent knowledge graphs with Gaussian embedding. Proceedings of the 24th ACM International Conference on Information and Knowledge Management, pp. 623-632, 2015.

[34]    H. Xiao, M. Huang, X. Zhu. From one point to a manifold: Knowledge graph embedding for precise link prediction. Proceedings of the 25th International Joint Conference on Artificial Intelligence, pp. 1315-1321, 2016.

[35]    T. Ebisu, R. Ichise. TorusE: Knowledge graph embedding on a lie group. Proceedings of the 32nd AAAI Conference on Artificial Intelligence (AAAI), pp. 1819-1826, 2018.

[36]    M. Nickel, V. Tresp, H. P. Kriegel. A three-way model for collective learning on multi-relational data. Proceedings of the 28th International Conference on Machine Learning, pp. 809-816, 2011.

[37]    B. Yang, W. T. Yih, X. He, et al. Embedding entities and relations for learning and inference in knowledge bases. Proceedings of the 3rd International Conference on Learning Representations, 2015.

[38]    I. Balazevic, C. Allen, T. Hospedales. TuckER: Tensor factorization for knowledge graph completion. Proceedings of the 2019 Conference on Empirical Methods in Natural Language Processing (EMNLP), pp. 5185-5194.

[39]    A. Bordes, X. Glorot, J. Weston, et al. A semantic matching energy function for learning with multi-relational data. Machine Learning, vol. 94, no. 2, pp. 233-259, 2014.

[40]    R. Socher, D. Chen, C. D. Manning, et al. Reasoning with neural tensor networks for knowledge base completion. Proceedings of the 26th International Conference on Neural Information Processing Systems, pp. 926-934, 2013.

[41]    T. Dettmers, P. Minervini, P. Stenetorp. Convolutional 2D knowledge graph embeddings. Proceedings of the 32nd AAAI Conference on Artificial Intelligence, pp. 1811-1818, 2018.

[42]    D. Q. Nguyen, T. D. Nguyen, D. Q. Nguyen, et al. A novel embedding model for knowledge base completion based on convolutional neural network. Proceedings of the 16th Annual Conference of the North American Chapter of the Association for Computational Linguistics, pp. 327-333, 2018.

[43]    D. Q. Nguyen, T. Vu, T. D. Nguyen, et al. A capsule network-based embedding model for knowledge graph completion and search personalization. Proceedings of the 2019 Conference of the North American Chapter of the Association for Computational Linguistics (NAACL), pp. 2180-2189.

[44]    S. Vashishth, S. Sanyal, V. Nitin, et al. InteractE: Improving convolution-based knowledge graph embeddings by increasing feature interactions. Proceedings of the 34th AAAI Conference on Artificial Intelligence, pp. 3009-3016, 2020.

[45]    T. Trouillon, J. Welbl, S. Riedel, et al. Complex embeddings for simple link prediction. Proceedings of the 33rd International Conference on Machine Learning, pp. 2071-2080, 2016.

[46]    M. Nickel, L. Rosasco, T. Poggio. Holographic embeddings of knowledge graphs. Proceedings of the 30th AAAI Conference on Artificial Intelligence, pp. 1955-1961, 2016.

[47]    S. M. Kazemi, D. Poole. SimplE embedding for link prediction in knowledge graphs. Proceedings of the 32nd International Conference on Neural

Information Processing Systems, pp. 4289-4300, 2018.

[48] L. Chao, J. He, T. Wang, et al. Pairre: Knowledge graph embeddings via paired relation vectors. Annual Meeting of the Association for Computational Linguistics and International Joint Conference on Natural Language Processing, pp. 4360-4369, 2021.

[49] L. Yu, Z. Luo, H. Liu, et al. Triplere: Knowledge graph embeddings via tripled relation vectors. arXiv preprint arXiv:2209.08271, 2022.

[50] X. Zhang, Q. Yang, D. Xu. Trans: Transition-based knowledge graph embedding with synthetic relation representation. arXiv preprint arXiv:2204.08301, 2022.

[51] Z. Sun, Z. H. Deng, J. Y. Nie, et al. RotatE: Knowledge graph embedding by relational rotation in complex space. Proceedings of the 7th International Conference on Learning Representations, pp. 1-18, 2019.

[52] S. Zhang, Y. Tay, L. Yao, et al. Quaternion knowledge graph embedding. Proceedings of the 33rd International Conference on Neural Information Processing Systems, pp. 2731-2741, 2019.

[53] Z. Cao, Q. Xu, Z. Yang, et al. Dual quaternion knowledge graph embeddings. Proceedings of the 35th AAAI Conference on Artificial Intelligence, pp. 6894-6902, 2021.

[54] J. Guo, S. Kok. Bique: Biquaternionic embeddings of knowledge graphs. Conference on Empirical Methods in Natural Language Processing, pp. 8338-8351, 2021.

[55] C. Xu, R. Li. Relation embedding with dihedral group in knowledge graph. Annual Meeting of the Association for Computational Linguistics, pp. 263-272, 2019.

[56] X. Ge, Y. C. Wang, B. Wang, et al. CompoundE: Knowledge graph embedding with translation, rotation and scaling compound operations. arXiv preprint arXiv:2207.05324, 2022.

[57] S. Wang, K. Fu, X. Sun, et al. Hierarchical-aware relation rotational knowledge graph embedding for link prediction. Neurocomputing, vol. 458, pp. 259-270, 2021.

[58] Y. Li, R. Zheng, T. Tian, et al. Joint embedding of hierarchical categories and entities for concept categorization and dataless classification. COLING, pp. 2678-2688, 2016.

[59] Z. Zhang, F. Zhuang, M. Qu, et al. Knowledge graph embedding with hierarchical relation structure. Conference on Empirical Methods in Natural Language Processing, pp. 3198-3207, 201

[60] R. Xie, Z. Liu, M. Sun. Representation learning of knowledge graphs with hierarchical types. Proceedings of the 25th International Joint Conference on Artificial Intelligence, pp. 2965-2971, 2016.

[61] M. Nickel, D. Kiela. Poincare embeddings for learning hierarchical representations. Proceedings of the 31st International Conference on Neural Information Processing Systems (NIPS), pp. 6341-6350, 2017.

[62] I. Balažević, C. Allen, T. Hospedales. Multi-relational Poincare graph embeddings. Proceedings of the 33rd International Conference on Neural Information Processing Systems, pp. 4463-4473, 2019.

[63] I. Chami, A. Wolf, D-C. Juan, et al. Low-dimensional hyperbolic knowledge graph embeddings. arXiv preprint arXiv:2005.00545, 2020.

[64] Z. Sun, M. Chen, W. Hu, et al. Knowledge association with hyperbolic knowledge graph embeddings. Conference on Empirical Methods in Natural Language Processing, pp. 5704-5716, 2020.

[65] J. Bruna, W. Zaremba, A. Szlam, et al. Spectral networks and locally connected networks on graphs. Proceedings of the 2nd International Conference on Learning Representations, 2014.

[66] S. Wang, X. Wei, C. N. Nogueira Dos Santos, et al. Mixed-curvature multi-relational graph neural network for knowledge graph completion. International Conference on World Wide Web, pp. 1761-1771, 2021.

[67] B. Xiong, S. Zhu, M. Nayyeri, et al. Ultrahyperbolic knowledge graph embeddings. arXiv preprint arXiv:2206.00449, 2022.

[68] Z. Zhang, J. Cai, Y. Zhang, et al. Learning hierarchy-aware knowledge graph embeddings for link prediction. AAAI Conference on Artificial Intelligence, pp. 3065-3072, 2020.

[69] S. Wang, X. Wei, C. N. Dos Santos, et al. Knowledge graph representation via hierarchical hyperbolic neural graph embedding. IEEE International Conference on Big Data, pp. 540-549, 2021.

[70] Z. Pan, P. Wang. Hyperbolic hierarchy-aware knowledge graph embedding for link prediction. Findings of Conference on Empirical Methods in Natural Language Processing, pp. 2941-2948, 2021.

[71] R. Xie, Z. Liu, J. Jia, et al. Representation learning of knowledge graphs with entity descriptions. Proceedings of the 30th AAAI Conference on Artificial Intelligence, pp. 2659-2665, 2016.

[72] L. Yao, C. Mao, Y. Luo. KG-BERT: BERT for knowledge graph completion. arXiv preprint arXiv:1909.03193, 2019.

[73] R. Xie, Z. Liu, H. Luan, et al. Image-embodied knowledge representation learning. Proceedings of the 26th International Joint Conference on Artificial Intelligence, pp. 3140-3146, 2017.

[74] M. Wang, S. Wang, H. Yang, et al. Is visual context really helpful for knowledge graph? A representation learning perspective. Proceedings of the 29th ACM International Conference on Multimedia, pp. 2735-2743, 2021.

[75] G. Niu, Y. Zhang, B. Li, et al. Rule-Guided Compositional Representation Learning on Knowledge Graphs. Proceedings of the AAAI Conference on Artificial Intelligence, pp. 2950-2958, 2020.

[76] G. Niu, B. Li, Y. Zhang, et al. Perform like an Engine: A Closed-Loop Neural-Symbolic Learning Framework for Knowledge Graph Inference. Proceedings of the 29th International Conference on Computational Linguistics, pp. 1391-1400, 2022.

[77] G. Niu, Y. Li, C. Tang, et al. Relational Learning with Gated and Attentive Neighbor Aggregator for Few-Shot Knowledge Graph Completion. Proceedings of the 44th International ACM SIGIR Conference on Research and Development in Information Retrieval, pp. 213-222, 2021.

[78] Z. Han, P. Chen, Y. Ma, et al. Dyernie: Dynamic evolution of Riemannian manifold embeddings for temporal knowledge graph completion. Proceedings of the 2020 Conference on Empirical Methods in Natural Language Processing, pp. 7301-7316.

[79] S. Ling, K. Nguyen, A. Roux-Langlois, et al. A lattice-based group signature scheme with verifier-local revocation. Theoretical Computer Science, vol. 730, no. 19, pp. 1-20, 2018.